\def\year{2020}\relax
\documentclass[letterpaper]{article} 
\usepackage{aaai20}  
\usepackage{times}  
\usepackage{helvet} 
\usepackage{courier}  
\usepackage[hyphens]{url}  
\usepackage{graphicx} 
\usepackage{graphicx}  
\usepackage{latexsym}
\usepackage{graphicx}
\usepackage{amsmath,amssymb}
\usepackage{booktabs}
\usepackage{longtable}
\usepackage{array}
\usepackage{multirow}
\usepackage{wrapfig}
\usepackage{float}
\usepackage{colortbl}
\usepackage{pdflscape}
\usepackage{comment}
\usepackage{tabu}
\usepackage{threeparttable}
\usepackage{threeparttablex}
\usepackage[normalem]{ulem}
\usepackage{makecell}
\usepackage{graphicx}
\usepackage{xcolor}
\usepackage{rotating}
\usepackage{url}
\usepackage{subcaption}
\usepackage{tabularx}
\usepackage[export]{adjustbox}
\usepackage{array}

\usepackage{bbm}
\newcommand{\citet}[1]{\citeauthor{#1}~\shortcite{#1}}

\title{Inducing Relational Knowledge from BERT}
\author{Zied Bouraoui\\
CRIL - CNRS \& Univ Artois, France\\
zied.bouraoui@cril.fr\\
\And
Jose Camacho-Collados\\
Cardiff University, UK\\
camachocolladosj@cardiff.ac.uk
\And
Steven Schockaert\\
Cardiff University, UK\\
schockaerts1@cardiff.ac.uk
}

\usepackage{graphicx}

\usepackage{url}
 
\begin{document}

\maketitle

\begin{abstract}
One of the most remarkable properties of word embeddings is the fact that they capture certain types of semantic and syntactic relationships. Recently, pre-trained language models such as BERT have achieved groundbreaking results across a wide range of Natural Language Processing tasks. However, it is unclear to what extent such models capture relational knowledge beyond what is already captured by standard word embeddings. To explore this question, we propose a methodology for distilling relational knowledge from a pre-trained language model. Starting from a few seed instances of a given relation, we first use a large text corpus to find sentences that are likely to express this relation. We then use a subset of these extracted sentences as templates. Finally, we fine-tune a language model to predict whether a given word pair is likely to be an instance of some relation, when given an instantiated template for that relation as input.
\end{abstract}

\section{Introduction}
Relation induction is the problem of predicting likely instances of a given relation based on some example instances of that relation. For instance, given the example pairs (\textit{paris}, \textit{france}), (\textit{tokyo}, \textit{japan}), (\textit{canberra}, \textit{australia}), a relation induction system should predict other instances of the capital-of relation (without explicitly being told that the relation of interest is the capital-of relation). By far the most common strategy is to treat this problem as a relation extraction problem. In such a case, sentences mentioning the example pairs are extracted from a large corpus and some neural network model is trained on these sentences. To predict new instances of the relation, the resulting model can then be applied to other sentences from the given corpus. 

One of the most surprising aspects of word embeddings, such as those learned using Skip-gram \cite{DBLP:journals/corr/abs-1301-3781} and GloVe \cite{glove2014}, is the fact that they capture relational knowledge, despite essentially being trained to capture word similarity. This is most clearly illustrated in the fact that predicting analogical word pairs is a commonly used benchmark for evaluating word embeddings. The problem of relation induction using word embeddings has also been studied \cite{Vylomova2016,DBLP:conf/coling/DrozdGM16,ziedcoling,vulic2018specialising,camacho-collados-etal-2019-relational}. In this case, new instances of the relation are predicted based only on pre-trained word vectors. Compared to relation extraction methods, the use of word vectors has the advantage that word pairs may be predicted even if they never co-occur in the same sentence, intuitively because they are sufficiently similar to the example pairs. Moreover, models that directly make predictions based on word vectors are much faster, among others because they do not have to retrieve relevant sentences from the corpus. On the other hand, relation induction methods based on word embeddings can be more noisy than those that rely on relation extraction.

Recently, the use of pre-trained language models such as BERT \cite{DBLP:conf/naacl/DevlinCLT19}, GPT-2 \cite{radford2019language}, and XLNet \cite{yang2019xlnet} has led to substantial performance increases in a variety of Natural Language Processing (NLP) tasks. A natural question is thus whether such language models capture more relational knowledge than standard word embeddings, and in particular whether they can lead to improved performance on the relation induction task. In particular, language models such as BERT and XLNet are trained to complete sentences containing blanks. By choosing sentences that express a relational property, we may thus be able to extract relational knowledge from these models. To explore this strategy, Table \ref{tabSentencesBERT} contains some predictions made by BERT for a number of different sentences\footnote{We experimented with XLNet as well, but its predictions were less accurate than those of BERT, possibly due to the short length of these test sentences.}. As can be seen, the performance is rather mixed. For example, this model does not seem to capture color properties, predicting either \textit{yellow} or \textit{white} for all examples, whereas it seems to have learned the capital-of relation well (notwithstanding the incorrect prediction for Brazil). The most important insight from Table \ref{tabSentencesBERT} comes from the two sentences about the cause of recessions, where the addition of the word \textit{often} makes a difference between a sensible prediction (\textit{inflation}) and a meaningless one (\textit{stress}). This suggests that even if language models capture relational knowledge, it is important to find the right sentences to extract that knowledge. 

In this paper, we propose a methodology for finding such trigger sentences based on a large text corpus. Similar as in relation extraction systems, we start by finding all sentences from the corpus that mention the example word pairs we have been given. We then filter these sentences to identify those that express the considered relation. To test whether a new word pair $(s,t)$ is an instance of the same relation, we then replace the example word pairs from the remaining sentences by the pair $(s,t)$ and use a language model (BERT in our experiments) to determine whether the resulting sentence is still natural. Crucially, note that the prediction about the pair $(s,t)$ does not rely on any sentences from the corpus mentioning $s$ and $t$. This means in particular that the accuracy of the predictions relies purely on the relational knowledge that is captured in the pre-trained language model. We only use the text corpus to find predictive trigger sentences.


\begin{table}
\resizebox{\columnwidth}{!}{
\begin{tabular}{ll}
\toprule
\textbf{Sentence} & \textbf{BERT} \\
\midrule
The color of the banana is \underline{\phantom{abc}}. &  yellow\\
The color of the avocado is \underline{\phantom{abc}}. &  yellow\\
The color of the carrot is \underline{\phantom{abc}}. &  yellow\\
The color of the tomato is \underline{\phantom{abc}}. &  white\\
The color of the kiwi is \underline{\phantom{abc}}. &  white\\
\midrule
The capital of Japan is \underline{\phantom{abc}}. & tokyo\\
The capital of France is \underline{\phantom{abc}}. & paris\\
The capital of Australia is \underline{\phantom{abc}}. & canberra\\
The capital of the US is \underline{\phantom{abc}}. & washington\\
The capital of Brazil is \underline{\phantom{abc}}. & santos\\
\midrule
Recessions are caused by \underline{\phantom{abc}}. & inflation\\
Recessions are often caused by \underline{\phantom{abc}}. & stress\\
Hangovers are caused by \underline{\phantom{abc}}. & stress\\
I took my umbrella because it was \underline{\phantom{abc}}. & warm\\
He didn't go to school because it was a \underline{\phantom{abc}}. & secret\\
\midrule
I like to have \underline{\phantom{abc}} for breakfast. & them\\
Her favorite subject in school was \underline{\phantom{abc}}. & english\\
His favorite day of the week is \underline{\phantom{abc}}. & christmas\\
They saw lots of scary animals such as \underline{\phantom{abc}}. & bears\\
He likes \underline{\phantom{abc}} and most other vegetables. & potatoes\\
\bottomrule
\end{tabular}}
\caption{Predictions by the BERT-Large-Uncased pre-trained language model for selected sentences. \label{tabSentencesBERT}}
\end{table}

\section{Related Work}
\textbf{Inducing knowledge from word embeddings.} Several authors have studied to what extent word embeddings capture meaningful attributional and relational knowledge. Most of these works are inspired by the finding of \citet{mikolov2013linguistic} that word embeddings capture analogies. 
For instance, \citet{DBLP:conf/acl/RubinsteinLSR15} analyzed how well pre-trained Skip-gram and GloVe vectors are able to predict properties of nouns, focusing on both taxonomic properties (e.g.\ being a bird) and attributive properties (e.g.\ being dangerous). In general, they obtained encouraging results for taxonomic properties but concluded that the ability of word vectors to predict attributive properties is limited. Similarly, \citet{gupta2015distributional} show that word vectors can to some extent predict ordinal attributes of cities and countries. For instance, they showed that countries can be ranked by GDP by training a linear regression model on the word vectors of the countries. The extent to which vector differences and other linear transformations between two words capture their relationship was also analyzed by subsequent works \cite{Vylomova2016,DBLP:conf/coling/DrozdGM16,ziedcoling}. 
While proved successful in many cases, and even somewhat supported by theoretical insights \cite{arora2016latent,DBLP:conf/icml/AllenH19}, simple linear transformations of word vectors have been found limiting in more general settings \cite{levy2014linguistic,linzen2016issues,rogers2017too,nissim2019analogies}.
Another line of work has therefore advocated to directly learn relation vectors from distributional statistics, i.e.\ vectors encoding the relationship between two words \cite{washio2018filling,DBLP:conf/acl/JameelSB18,DBLP:conf/coling/AnkeS18,joshi2018pair2vec,DBLP:conf/emnlp/WashioK18,relative2019ijcai}. 

\smallskip
\noindent \textbf{Inducing knowledge from language models.} Recently, probing tasks have been used to better understand the nature of the representations learned by neural language models, although most works have generally focused on linguistic aspects \cite{DBLP:conf/naacl/HewittM19,DBLP:journals/corr/abs-1901-05287,DBLP:conf/acl/JawaharSS19,DBLP:conf/acl/TenneyDP19}. More closely related to our work, \citet{forbes2019neural} analyze to what extent properties and affordances of objects can be predicted using neural language models, by relying on manually chosen sentences. For instance, to determine whether accordions are squishy, they consider the sentence ``An accordion is squishy''. Whether the property applies or not is then predicted from the resulting output of the language model (e.g.\ the output vector of the [CLS] token in the case of BERT). When training this classifier, they also fine-tune the pre-trained language model. Our work differs in that we consider arbitrary relations (as opposed to the object-property and object-affordance relations) and the fact that we automatically identify the most appropriate trigger sentences for each relation. The problem of extracting relational knowledge from the BERT language model was also studied very recently in \citet{langmodelknowledgebases2019}. In this work, a wide range of relations is considered, but their approach again depends on manually chosen trigger sentences. Another difference with our work is that they focus on predicting tail words $t$ that are related to a given source word $s$, whereas we focus on relation classification. Finally, \citet{bosselut-etal-2019-comet} propose an automatic knowledge graph construction method based on transformer language models. However, they rely on large amounts of training data\footnote{100K triples for learning 34 relation types for their ConceptNet experiments, and 710K training triples for ATOMIC.}, whereas we focus on settings where only a handful of training examples are given, thereby relying on the relational knowledge captured by BERT instead of the ability of the model to generalize. 

Within a broader context, the importance of finding the right input sentences when extracting knowledge from language models was also considered by \citet{DBLP:conf/emnlp/AmramiG18}. Specifically, they consider the problem of characterizing word senses using language models. For instance to characterize the sense of the word \textit{sound} in the sentence ``I liked the sound of the harpsichord'', a standard approach would be to look at the predictions of a language model for the input ``I liked the \underline{\phantom{abc}} of the harpsichord''. However, they found that better results can be obtained by instead considering the sentence ``I liked the sound and \underline{\phantom{abc}} of the harpsichord''. Finally, \citet{DBLP:conf/acl/LoganLPGS19} have pointed out that neural language models are severely limited in their ability to capture factual knowledge, which they use as a motivation to study knowledge graph enhanced language models.


\section{Methodology}
In this section we describe a method for relation induction using language models. As highlighted in the introduction, a key challenge is to find the right linguistic pattern to use as input to the language model. Let us write $\phi(h,t)$ to denote a sentence which mentions some head word $h$ and tail word $t$. For instance, consider the following sentence:
$$
\phi(\textit{Paris},\textit{France}) = \textit{\underline{Paris} is the capital of \underline{France}.}
$$
We will treat such sentences as templates, which can be instantiated with different word pairs, e.g.:
\begin{align*}
\phi(\textit{Rome},\textit{Italy}) &= \textit{\underline{Rome} is the capital of \underline{Italy}.}\\
\phi(\textit{Rome},\textit{France}) &= \textit{\underline{Rome} is the capital of \underline{France}.}\\
\phi(\textit{Trump},\textit{Obama}) &= \textit{\underline{Trump} is the capital of \underline{Obama}.}
\end{align*}
Our main intuition is that a language model should be able to recognize that the sentence $\phi(\textit{Rome},\textit{Italy})$ is natural, while $\phi(\textit{Rome},\textit{France})$ and $\phi(\textit{Trump},\textit{Obama})$ are not. Note, however, that this example relies on the fact that the template $\phi$ is indicative of the capital-of relation. Clearly this is not the case for all sentences mentioning \textit{Paris} and \textit{France}. For instance, consider the following sentence:
$$
\phi'(\textit{Paris},\textit{France}) = \textit{The Eiffel tower is in \underline{Paris}, \underline{France}.}
$$
A sentence such as $\phi'(\textit{Rome},\textit{Italy})$ is clearly not natural, hence we cannot use $\phi'$ to find new instances of the capital-of relation. 
Let us assume that the following examples of a given relation are given: $\{(s_1,t_1),...,(s_n,t_n)\}$. Based on the aforementioned intuitions, we propose a strategy for finding likely additional instances of that relation, consisting of the following three steps:
\begin{enumerate}
\item Find all sentences $\phi(s_i,t_i)$ mentioning the source and target word of one of the given examples. 
\item Filter the resulting templates $\phi$, keeping only those which seem to express the considered relationship. In particular, to determine the adequacy of a template $\phi$, we check whether a pre-trained language model can predict the corresponding tail words $t_1,...,t_n$ from the sentences $\phi(s_1,\underline{\phantom{a}}),...,\phi(s_n,\underline{\phantom{a}})$ and the corresponding head words $h_1,...,h_n$ from the sentences $\phi(\underline{\phantom{a}},t_1),...,\phi(\underline{\phantom{a}},t_n)$.
\item Fine-tune a language model to predict from instantiations $\phi(s,t)$ of the remaining templates whether $(s,t)$ is likely to be an instance of the relation.
\end{enumerate}
We now explain these steps in more detail.


\subsection{Finding Candidate Sentences}
The first step is straightforward. Given a set of word pairs $\mathcal{R} = \{(s_1,t_1),...,(s_n,t_n)\}$, we extract all sentences mentioning one of these word pairs $(s_i,t_i)$. We will use a Wikipedia corpus for this purpose, although other large corpora would also be suitable. We only consider sentences with at most 100 words and a maximum window size of 15 between the occurrences of the words $s_i$ and $t_i$.

\subsection{Filtering Templates}
\label{filtering}

Let $\phi_1(x_1,y_1),...,\phi_m(x_m,y_m)$ be the set of all sentences extracted for the given set of word pairs $\mathcal{R}$, where $(x_i,y_i) \in \mathcal{R}$ for every $i$. The aim of the filtering step is to select templates $\phi_j$ which are such that most of the sentences $\phi_j(s_1,t_1),...,\phi_j(s_n,t_n)$ are natural. In other words, we want to identify sentences $\phi_j(x_j,y_j)$ which express the considered relationship in general, rather than being specifically about $x_j$ and $y_j$. For many of the extracted sentences this may not be the case, as they might simply mention the two words for an unrelated reason (e.g.\ ``\underline{Paris} Hilton arrived in \underline{France} today.'') or they might only be sensible for the particular word pair (e.g.\ ``The Eiffel Tower is located in \underline{Paris}, \underline{France}.''). Moreover, some sentences may not directly express the considered relationship, but might nonetheless provide some useful evidence. Consider for instance the following sentences:
\begin{align}
\phi_1&:\quad \textit{\underline{Paris} is located in central \underline{France}.}\label{eqSentIndirect1}\\
\phi_2&:\quad\textit{\underline{Paris} is the largest city in \underline{France}.}\\
\phi_3&:\quad\textit{\underline{Paris} is one of the oldest cities in \underline{France}.}\label{eqSentIndirect3}
\end{align}
While none of these sentences asserts the capital-of relationship, a word pair $(s,t)$ for which the assertions $\phi_1(s,t)$, $\phi_2(s,t)$ and $\phi_3(s,t)$ are all true is nonetheless likely to be an instance of the capital-of relation. The problem we consider is thus to rank the templates $\phi_1,...,\phi_m$ by their usefulness. If there are any templates that directly express the relation, then those should ideally be used. However, for many commonsense relations, we may not have such sentences as commonsense knowledge is rarely asserted explicitly \cite{DBLP:conf/cikm/GordonD13}, in which case we have to instead rely on sentences providing indirect evidence, such as \eqref{eqSentIndirect1}--\eqref{eqSentIndirect3}.

To assess the usefulness of the template $\phi_i$, we use a pre-trained BERT model to fill in the blanks in the sentences $\phi_i(s_1,\underline{\phantom{a}}),...,\phi_i(s_n,\underline{\phantom{a}})$ and $\phi_i(\underline{\phantom{a}},t_1),...,\phi_i(\underline{\phantom{a}},t_n)$, where we write e.g.\ $\phi_i(s_1,\underline{\phantom{a}})$ for the sentence $\phi_i(s_1,t_1)$ in which $t_1$ was replaced by a blank. Note that other masked language models such as XLNet could also be used. We then simply count for how many of these $2n$ sentences the correct word was among the top-$k$ predictions. Specifically, let us write $T_{ij}$ for the set of top-$k$ predictions for the sentence $\phi_i(s_j,\underline{\phantom{a}})$ and $S_{ij}$ for the set of top-$k$ predictions for the sentence $\phi_i(\underline{\phantom{a}},t_j)$. The templates $\phi_i$ are then ranked based on the following score:
\begin{align}\label{eqScoreSlow}
\textit{score}_1(\phi_i) = \sum_{j=1}^n \mathbbm{1}[s_j \in S_{ij}] + \mathbbm{1}[t_j \in T_{ij}]
\end{align}
where $\mathbbm{1}[s_j \in S_{ij}]$ is 1 if $s_j \in S_{ij}$ holds and 0 otherwise, and similar for $\mathbbm{1}[t_j \in T_{ij}]$.

Given the large number of extracted sentences and the possibly large set of pairs in $\mathcal{R}$, applying this score to all sentences would be prohibitively expensive. Therefore, we first select a subset of the templates $\phi_1,...,\phi_m$ based on a faster scoring function. In particular, for each sentence $\phi_i(x_i,y_i)$, we use the language model to obtain the top-k predictions $T_i$ for the variant $\phi_i(x_i,\underline{\phantom{a}})$, and the top-k predictions $S_i$ for the variant $\phi_i(\underline{\phantom{a}},y_i)$. Then we use the following score:
\begin{align}\label{eqScoreFast}
\textit{score}_2(\phi_i) = |S_i \cap \{s_1,...,s_n\}| + |T_i \cap \{t_1,...,t_n\}|
\end{align}
Note that with this score, we only need to make two top-$k$ predictions for each of the sentences, whereas \eqref{eqScoreSlow} requires us to make $2n$ predictions for each sentence. The score \eqref{eqScoreFast} intuitively checks whether most of the top-$k$ predictions are of the correct type. In other words, even if the predictions made by the language model are wrong, if they are at least of the correct type (e.g.\ the name of a country, if we are predicting the tail word of a capital-of relation), we can have some confidence that the template is meaningful.



\subsection{Fine-tuning BERT}
Let us write $\psi_1,...,\psi_k$ for the templates that were selected after the filtering step. It is straightforward to use these templates for link prediction, which is the task of finding a tail word $t$, given some source word $s$, such that $(s,t)$ is an instance of the considered relation. Indeed, to find plausible tail words $t$, we can simply aggregate the predictions that are made by a masked language model for the sentences $\phi_1(s,\underline{\phantom{a}}),...,\phi_k(s,\underline{\phantom{a}})$. Our main focus, however, is on relation induction. More specifically, given a candidate pair $(s,t)$ we consider the problem of determining whether $(s,t)$ is likely to be a correct instance of the considered relation. In this case, it is not sufficient that $t$ is predicted for some sentence $\phi_i(s,\underline{\phantom{a}})$. To illustrate this, consider the following non-sensical instantiation of a capital-of template:
$$
\textit{The capital of Macintosh is \underline{\phantom{abc}}.}
$$
One of the top predictions\footnote{The only two higher ranked words were \textit{macintosh} and \textit{mac}.} by the BERT-Large-Uncased model is \textit{Apple}, which might lead us to conclude that (\textit{Macintosh}, \textit{Apple}) is an instance of the capital-of relation.

Rather than trying to classify a given word pair $(s,t)$ by filling in blanks, we will therefore use the full sentence $\phi(s,t)$ as input to the BERT language model, and train a classifier on top of the output produced by BERT. In particular, we use the output vector for the [CLS] token, which has been shown to capture the overall meaning of the sentence \cite{DBLP:conf/naacl/DevlinCLT19}. Our hypothesis is that the vector $h_{[CLS]}$ which is predicted for the [CLS] token will capture whether the input sentence is natural or unusual, and thus whether $(s,t)$ is likely to be a valid instance of the relation. In particular, we add a classification layer that takes the $h_{[CLS]}$ vector as input and predicts whether the input sentence $\phi_i(s,t)$ is a correct assertion, i.e.\ whether the pair $(s,t)$ is an instance of the considered relation. Note that a single classifier is trained for each given relation (i.e.\ regardless of which template $\phi_i$ was used to construct the input sentence). Since the way in which we use the output from BERT is different from how it was trained, we fine-tune the parameters of BERT while training the classification layer.

The given set $\mathcal{R}$ contains positive examples of word pairs that have the considered relation. However, to train the classifier we also need negative examples. To this end, following \citet{Vylomova2016}, we follow two strategies for corrupting the examples from $\mathcal{R}$. First, for an instance $(s,t)\in \mathcal{R}$ we use $(t,s)$ as a negative example (provided that  $(t,s)\notin \mathcal{R}$). Second, we also construct negative examples of the form $(s_i,t_j)$ by combining the source word of one pair from $\mathcal{R}$ with the tail word of another pair. 

For the classification layer, we use a linear activation function, with a binary cross-entropy loss. To optimize the loss function, we uses Adam with fixed weight decay and warmup linear schedule.


\subsection{Relation Classification}
Given a word pair $(s,t)$, we obtain $k$ predictions about whether this pair is likely to be an instance of the relation, i.e.\ one prediction for each considered template. Let us write the corresponding probabilities as $p_1(s,t),...,p_k(s,t)$. To combine these predictions, we consider two strategies. 
With the first strategy, we predict $(s,t)$ to be a positive example if $\max_i p_i(s,t) > 1- \min_i p_i(s,t)$. We will refer to this model as \textsc{BERT}$^{\max}$. In other words, in this case we check whether there is a positive prediction which has higher confidence than any of the negative predictions. For the second strategy, we instead look at an average (or sum) across all templates. In particular, we then predict $(s,t)$ to be positive if $\sum_i p_i(s,t) \geq \lambda$, with $\lambda$ a threshold which is selected based on held-out tuning data. We will refer to this model as \textsc{BERT}$^{*}$.


\section{Experiments}
In this section, we experimentally analyze the performance of our method. Our main question of interest is whether the proposed method allows us to model relations in a better way than is possible with pre-trained word vectors.

\begin{table*}
\centering
\begin{tabular}{|l || ccc |  ccc | ccc | }
\hline
& \multicolumn{3}{c|}{Google} & \multicolumn{3}{c|}{DiffVec} & \multicolumn{3}{c|}{BATS}  \\
      
& pr & rec & f1     & pr & rec & f1    & pr & rec & f1    \\
\hline
SVM\textsubscript{glove}   &45.7& 70.2 & 55.3                & 32.7 & 52.7 & 40.3   & 42.3 & 55.6 & 48.0                \\
SVM\textsubscript{sg}   & 49.4 & 68.9 & 57.5  &    38.5 & 47.2 & 42.4                  &   42.9 & 61.3 & 50.4              \\
\hline
Trans\textsubscript{glove} &  76.9 & 72.5 & 74.6    &  39.6&59.6&47.5                &   53.4& 65.6 & 58.8                \\
Trans\textsubscript{sg}  &  73.1 & \textbf{74.3} & 73.6    &   47.3 & \textbf{72.6} & \textbf{57.2}                 &    \textbf{63.1} & \textbf{70.6} & \textbf{66.6}              \\
\hline
\hline 
BERT$^{\max}$\textsubscript{50}  &  85.2 & 67.1 & 75.0    &  58.1 & 43.4 & 49.6      &  57.3 & 36.5& 44.5           \\
BERT$^{\max}$\textsubscript{100}  & \textbf{86.8} & 69.3 & \textbf{77.0}    &  59.5 & 46.7 & 52.8      &  60.3 & 41.7 & 49.5               \\
BERT$^{\max}$\textsubscript{1000}  & 75.8 & 58.2 & 65.8      &  52.9 & 40.3 & 45.7   &  56.3 & 37.1 & 44.7       \\
\hline
BERT$^*$\textsubscript{50}  &  78.6 & 61.8 & 69.1                              &  51.1 & 39.2 & 44.3        &   50.3 & 32.4 & 39.4           \\
BERT$^*$\textsubscript{100}  &  79.4 & 63.7 & 70.6                   &  \textbf{63.2} & 47.8 & 54.4       &   59.2 & 44.5 & 50.8          \\
BERT$^*$\textsubscript{1000}  &   76.9 & 51.0 & 61.3                               &   53.1 & 38.5 & 44.6    &  57.6 & 35.3 & 43.7   \\
\hline
\end{tabular}
\caption{Overview of the experimental results. \label{tabOverview}}
\end{table*}

\subsection{Experimental Setting}
\noindent\textbf{Benchmark datasets.} We consider relations taken from the following three standard benchmark datasets: 
\begin{itemize}
\item the Google analogy Test Set (Google), which contains 14 types of relations with a varying number of instances per relation \cite{DBLP:journals/corr/abs-1301-3781};
\item the Bigger Analogy Test Set (BATS), which contains 40 relations with 50 instances per relation \cite{DBLP:conf/naacl/GladkovaDM16};
\item the DiffVec Test Set (DV) contains 36 relations with a varying number of instances per relation \cite{Vylomova2016}.
\end{itemize}
Note that while these datasets contain both syntactic and semantic relationships, we can expect that the proposed method is mostly tailored towards semantic relationships.

\smallskip
\noindent\textbf{Experimental design.} For all datasets, we consider the corresponding relations in isolation, i.e.\ we model the relation induction task as a binary classification problem. To this end, for a given relation, we first split the set of available examples in two sets: a training set that contains 90\% of words pairs and a test set that contains the remaining 10\%. We use the examples from the training set to find relevant sentences (i.e., sentences where these word pairs in the relation co-occur) from the English Wikipedia corpus\footnote{We used the dump of May 2016.}. These sentences are filtered to find predictive patterns, and to train the classifiers and fine-tune the BERT-Large-Uncased language model\footnote{We used the BERT implementation available at \url{https://github.com/huggingface/transformers}}. To filter the set of templates, we first select the top 1000 templates using \eqref{eqScoreFast}, for each considered relation. We then select the $K$ most promising templates among them, using \eqref{eqScoreSlow}. We will separately show results for $K=50$, $K=100$ and $K=1000$. 

The test set is used to evaluate the model. Note that the test set only contains positive examples. To generate negative test examples, we follow the strategies proposed by \citet{Vylomova2016}. First, we consider the two strategies that we also used for generating negative examples for training the classifiers. In particular, for each pair $(s,t)$ in the test set, we add $(t,s)$ as a negative example, and for each source word $s$ in the test set, we randomly sample two target words from the test set (provided that the test set contains enough pairs), each time verifying that the generated negative examples do not in fact occur as positive examples. Furthermore, for each positive example, we also randomly select an instance from one of the other relations. Finally, for each positive example, we generate one random word pair from the set of all words that occur in the dataset. This ensures that the evaluation involves negative examples that consist of related words as well as negative examples that consist of unrelated words. Note that the number of negative examples is thus five times higher than the number of positive examples, which makes the task quite challenging.

\smallskip
\noindent\textbf{Baselines.} The use of word vector differences is a common choice for modelling relations using pre-trained word embeddings. As a first baseline we will consider a linear SVM classifier, but with more informative features than the vector difference. In particular, following \citet{vu2018integrating} we will represent a given word pair $(s,t)$ as $\mathbf{s}\oplus \mathbf{t} \oplus (\mathbf{s}\odot \mathbf{t})$, where we write $\oplus$ for vector concatenation, $\mathbf{s}$ and  $\mathbf{t}$ are the vector representations of $s$ and $t$, and we write $\mathbf{s}\odot \mathbf{t}$ for the component-wise product of $\mathbf{s}$ and $\mathbf{t}$. 
As a second baseline, we will use the model from \citet{ziedcoling}, which learns a Gaussian distribution over vector differences that are likely to correspond to word pairs from the considered relation. This distribution is combined with two other Gaussian distributions, which respectively capture the distribution of words that are likely to appear as source words (in valid instances of the relation) and the distribution of words that are likely to appear as target words. We refer to this baseline as \textit{Trans}. It was shown in \cite{ziedcoling} to outperform SVM classifiers trained on the vector difference. Note that while the SVM baseline uses the same positive and negative examples for training as our model, the \textit{Trans} baseline is a generative model which only uses the positive examples.
%
%

\smallskip
\noindent\textbf{Word representation.} As static word embeddings for the baselines, we will use the Skip-gram word vectors that were pre-trained from the 100B words Google News data set\footnote{\url{https://code.google.com/archive/p/word2vec/}} (SG-GN) and GloVe word vectors which were pre-trained from the 840B words Common Crawl data set\footnote{\url{https://nlp.stanford.edu/projects/glove/}} (GloVe-CC).

\begin{table}
\footnotesize
\centering
\begin{tabular}{| @{\hspace{3pt}}c@{\hspace{3pt}} |l || @{\hspace{5pt}}c@{\hspace{5pt}}c@{\hspace{5pt}}c@{\hspace{5pt}}|}
\hline
& Google                             & Trans\textsubscript{sg}    & SVM\textsubscript{sg}    & BERT$^{\max}_{100}$ \\
\hline
\multirow{9}{*}{\rotatebox{90}{Morphological}} 
& gram1-adj-to-adv           &   \textbf{63.5}    &   51.2    &  49.9  \\
& gram2-opposite                      &   59.2  &   49.6    &    \textbf{68.5} \\
& gram3-comparative                   &   \textbf{79.7}    &  62.1    &  78.4  \\
& gram4-superlative                   &   \textbf{88.3}    &   49.4    &  86.6   \\
& gram5-present-participle            &   \textbf{70.1}    &   56.1    &  68.9 \\
& gram6-nationality-adj         &   63.8     &   58.3    &  \textbf{79.6}\\       
& gram7-past-tense                    &   \textbf{80.1}     &  54.2   &   67.6  \\
& gram8-plural                        &   \textbf{72.9}     &   68.9    &  48.8   \\
& gram9-plural-verbs                  &   \textbf{69.4}    &   51.1    &  65.8  \\
\hline
\multirow{5}{*}{\rotatebox{90}{Semantic}} 
& currency                            &   82.3     &   60.1    &  \textbf{93.6}   \\
& capital-common-countries            &   82.3     &   73.4    & \textbf{91.2}   \\
& capital-world                       &   78.1     &   62.0    &  \textbf{89.5}\\
& family                              &   72.7     &   52.3    & \textbf{88.2}    \\
& city-in-state                       &   68.4     &   57.2    &  \textbf{79.6} \\
\hline
\end{tabular}
\caption{Breakdown of results for the Google analogy dataset (F1).}
\label{tabResultsGoogle}
\end{table}

\begin{table}[t]
\footnotesize
\centering
\begin{tabular}{|@{\hspace{3pt}}c@{\hspace{3pt}}|l@{\hspace{1pt}} || @{\hspace{3pt}}c@{\hspace{3pt}}c@{\hspace{3pt}}c@{\hspace{1pt}}|}
\hline
& DiffVec                        & Trans\textsubscript{sg}    & SVM\textsubscript{sg}    & BERT$^{\max}_{100}$ \\
 \hline
\multirow{3}{*}{\rotatebox{90}{Attr.\ }} 
& Action:ObjectAttribute         &   19.2     &  20.1     & \textbf{35.2}    \\
& Object:State                 &  56.2    &   32.1    & \textbf{58.0}    \\
& Object:TypicalAction           & 25.3 & 35.4 & \textbf{49.0}     \\
\hline
\multirow{8}{*}{\rotatebox{90}{Causality}}
& Action/Activity:Goal              & 31.9 & 29.3 & \textbf{57.1}    \\
& Agent:Goal                     & 43.5 & 36.7 & \textbf{53.9}     \\
& Cause:CompensatoryAction        & 59.1 & 46.8 & \textbf{63.4}    \\
& Cause:Effect                    & 63.4 & 42.4 & \textbf{64.0}     \\
& EnablingAgent:Object         & 34.3 & 45.5 & \textbf{58.7}      \\
& Instrument:Goal               & 56.8 & 41.2 & \textbf{60.5}     \\
& Instrument:IntendedAction   & 62.9 & 39.2 & \textbf{68.8}    \\
& Prevention                 & 70.1 & 53.2 & \textbf{72.1}       \\
\hline
\multirow{4}{*}{\rotatebox{90}{Lexical}} 
& Collective noun         & \textbf{55.6} & 40.8 & 38.1      \\
& Hyper                   & \textbf{73.6} & 41.5 & 54.3    \\
& Lvc                   & 75.0 & \textbf{75.6} & 37.4    \\
& Mero                 & \textbf{64.6} & 41.4 & 47.5     \\
\hline
\multirow{15}{*}{\rotatebox{90}{Commonsense}}
& Event                    & 50.2 & 39.8 & \textbf{57.8}        \\
& Concealment             & 42.1 & 32.4 & \textbf{52.7}     \\
& Expression               & \textbf{80.3} & 52.3 & 79.3          \\
& Knowledge                 & 70.1 & 51.4 & \textbf{72.4}         \\
& Plan                       & 56.5 & 32.3 & \textbf{62.3}        \\
& Representation           & 48.2 & 39.7 & \textbf{50.1}         \\
& Sign:Significant        & 38.1 & 30.2 & \textbf{41.1}        \\
& Attachment           & 36.4 & 41.0 & \textbf{52.9}      \\
& Contiguity              & 61.2 & 32.8 & \textbf{70.8}          \\
& Item:Location           & 28.1 & 32.1 & \textbf{54.2}         \\
& Loc:Action/Activity      & 74.8 & 51.3 & \textbf{77.4}       \\
& Loc:Instr/AssociatedItem  & 42.0 & 44.9 & \textbf{69.0}      \\
& Loc:Process/Product                 & 47.2 & 56.6 & \textbf{64.3}    \\
& Sequence                           & 62.8 & 50.2 & \textbf{74.9}       \\
& Time:Action/Activity            & 57.2 & 53.7 & \textbf{59.1}      \\
\hline
\multirow{6}{*}{\rotatebox{90}{Morphological}} 
& Noun Singplur               & \textbf{53.0} & 38.5 & 33.5   \\
& Prefix re                & \textbf{71.5} & 30.2 & 19.6     \\
& Verb 3rd                 & \textbf{97.0} & 38.4 & 20.3         \\
& Verb 3rd Past           & \textbf{95.3} & 32.2 & 21.9          \\
& Verb Past               & \textbf{82.1} & 61.3 & 26.6         \\
& Vn-Deriv               & \textbf{75.5} & 63.1 & 25.0         \\
\hline
\end{tabular}
\caption{Breakdown of results for the DiffVec dataset (F1).}
\label{tabResultsDiffVec}
\end{table}

\subsection{Results}

An overview of the results is presented in Table \ref{tabOverview}. In this table, for our model, we show results for three different values of $K$ (i.e.\ the number of selected templates after filtering), which are indicated in subscript. We can see that there are no consistent differences between the BERT$^{\max}$ and BERT$^*$ variants, and that the choices $K=100$ outperforms $K=50$ and $K=1000$. Note that the choice $K=1000$ corresponds to a setting where the scoring function \eqref{eqScoreFast} is not used. The weaker performance for that setting thus clearly shows the usefulness of our proposed scoring function. When comparing the results to the baselines, we can see that our model does not consistently outperform the \textit{Trans} baseline. This is most notable in the case of BATS, where \textit{Trans} performs overall much better. However, this is not unexpected given that these datasets contain a large number of morphological relationships (often also referred to as syntactic relations in this context), and there is not reason to expect why our proposed method should be able to perform well on such relations. For instance, we are unlikely to find many meaningful templates which express that $t$ is the plural of $s$.

\begin{table*}[t]
\centering
\setlength{\tabcolsep}{3.0pt}
\renewcommand\thetable{6}
\renewcommand{\arraystretch}{1.2}
\resizebox{\textwidth}{!}{
\begin{tabular}{|c|c|}
\hline
\multicolumn{1}{|c|}{\large \textbf{Currency}} 
& \multicolumn{1}{c|}{\large \textbf{Capital-of}} \\
\hline
Sales of all products and services traded online in * in 2012 counted 311.6 billion *
 &  Summer olympics, which were in *, the capital of the home country, * \\
 As is often the case in *, lottery ticket prices above the 80 * threshold are negotiable
 &  The main international airport serves *, the capital of and most populous city in * \\
The Government of * donated 300 million * to finance the school's construction in 1975
 &  It is located in *, the capital of * \\
 On his return to *, he had made 18,000 * on an initial investment of 4,500
 &  In 2006, he portrayed John Morton on a tour of * arranged by the US Embassy in * \\
 The cost of vertebroplasty in * as of 2010 was 2,500 *
 &  At the time, Jefferson was residing in *, while serving as American Minister to * \\
\hline
\end{tabular}
}
\caption{Automatically-extracted templates filtered by BERT associated with the \textit{currency} and \textit{capital-of} relations from the Google analogy dataset. \label{templatesqualitative1}}
\end{table*}

\begin{table}[h!]
\footnotesize
\centering
\renewcommand\thetable{5}
\begin{tabular}{|@{\hspace{3pt}}c@{\hspace{3pt}}|l@{\hspace{3pt}} || @{\hspace{3pt}}c@{\hspace{3pt}}c@{\hspace{3pt}}c@{\hspace{1pt}}|}
\hline
&BATS                        & Trans\textsubscript{sg}    & SVM\textsubscript{sg}    & BERT$^{\max}_{100}$ \\
 \hline
 \multirow{20}{*}{\rotatebox{90}{Morphological}} 
 & Regular plurals                   & \textbf{76.3} & 40.8 & 35.0    \\
 &Plurals - orth.\ changes        & \textbf{76.0} & 48.1 & 25.5     \\
 &Comparative degree               & \textbf{76.2} & 47.5 & 50.2    \\
 &Superlative degree               & \textbf{82.1} & 59.5 & 53.3  \\
 &Infinitive: 3Ps.Sg               & \textbf{82.0} & 59.8 & 25.5 \\ 
 &Infinitive: participle          & \textbf{79.4} & 62.7 & 33.3    \\
 &Infinitive: past              & \textbf{70.9} & 52.0 & 35.1    \\
 &Participle: 3Ps.Sg            & \textbf{78.3} & 62.9 & 29.9    \\
 &Participle: past               & \textbf{76.3} & 56.7 & 36.7     \\
 &3Ps.Sg: past                    & \textbf{86.4} & 65.8 & 25.9   \\

 &Noun+less                       & \textbf{62.5} & 43.8 & 26.6   \\
 &Un+adj                         & \textbf{71.2} & 40.5 & 28.8  \\
 &Adj+ly                        & \textbf{73.0} & 39.8 & 35.5 \\
 &Over+adh./Ved                  & \textbf{71.1} & 41.5 & 36.7    \\
 &Adj+ness                        & \textbf{72.5} & 53.6 & 30.5    \\
 &Re+verb                       & \textbf{75.1} & 56.8 & 33.9    \\
 &Verb+able                     & \textbf{73.8} & 55.3 & 25.4    \\
 &Verb+er                        & \textbf{60.2} & 53.3 & 42.3    \\
 &Verb+ation                      & \textbf{58.9} & 46.6 & 28.8  \\
 &Verb+ment                        & \textbf{60.6} & 48.1 & 40.7    \\
\hline
\multirow{10}{*}{\rotatebox{90}{Lexical}}
&Hypernyms animals                   & 63.6 & 64.5 & \textbf{71.8}    \\
& Hypernyms misc                    & 78.1 & 56.2 & \textbf{78.8}     \\
& Hyponyms misc                     & 54.6 & 50.9 & \textbf{61.3}    \\
& Meronyms substance                 & \textbf{53.1} & 37.8 & 50.4   \\
& Meronyms member                    & \textbf{70.2} & 57.1 & 56.6  \\
& Meronyms part-whole               & 49.5 & 52.3 & \textbf{58.2}   \\
& Synonyms intensity                & 46.7 & 35.6 & \textbf{50.8}    \\
& Synonyms exact                    & 41.3 & 29.9 & \textbf{48.7}    \\
& Antonyms gradable                 & \textbf{49.3} & 51.9 & 48.5    \\
& Antonyms binary                    & 49.6 & 33.3 & \textbf{54.5}   \\
\hline
\multirow{10}{*}{\rotatebox{90}{Encyclopedic}}
& Capitals                           & 68.6 & 52.1 & \textbf{73.2}    \\
& Country:language                   & 62.8 & 53.5 & \textbf{69.5}    \\
& UK city: county                    & 61.6 & 48 & \textbf{71.8}   \\
& Nationalities                       & 83.3 & 61.5 & \textbf{84.4} \\
& Occupation                         & 61.8 & 49.9 & \textbf{72.6}    \\
& Animals young                       & 51.2 & 50.7 & \textbf{68.2}    \\
& Animals sounds                      & 60.1 & 45.9 & \textbf{63.1} \\
& Animals shelter                    & 45.8 & 45.2 & \textbf{63.3}    \\
& thing:color                         & 75.6 & 58.9 & \textbf{76.5}     \\
& male:female                      & 76.9 & 49.3 & \textbf{79.0}     \\
 \hline
\end{tabular}
\caption{Breakdown of results for the BATS dataset (F1).}
\label{tabResultsBATS}
\end{table}

\begin{table}
\centering
\setlength{\tabcolsep}{3.0pt}
\renewcommand{\arraystretch}{1.2}
\resizebox{\columnwidth}{!}{
\begin{tabular}{|c|c|c|l|l|}
\hline
\multicolumn{1}{|c|}{\multirow{2}{*}{\textbf{Dataset}}} 
& \multicolumn{1}{c|}{\multirow{2}{*}{\textbf{Manual template}}} 
& \multicolumn{1}{c|}{\multirow{2}{*}{\textbf{Relation}}} & \multicolumn{2}{c|}{\textbf{Score}}  \\
& & & \multicolumn{1}{c}{\textbf{Hand}} & \multicolumn{1}{c|}{\textbf{Auto}}    \\
\hline
\multirow{2}{*}{Google} & \textit{* is the capital of *} & capital-world & 51.4 & 89.5   \\
 & \textit{* is the currency of *} & currency & 46.2 & 93.6  \\
 \hline
\multirow{2}{*}{DiffVec} & \textit{* is found in the *} & item:location & 28.6 & 54.2   \\
 & \textit{* is used to conceal *} & concealment & 25.7 & 52.7  \\
\hline
\end{tabular}
}
\caption{Hand-crafted templates associated with specific relations and their F1 scores. 
The score obtained with our automatic pipeline is shown as reference under "Auto". \label{templatesqualitative2}}
\end{table}

Therefore, in Tables \ref{tabResultsGoogle}--\ref{tabResultsBATS}, we compare the performance for the individual relations contained in the three datasets. In this case, we only show results for the BERT$^{\max}_{100}$ variant of our model. The results on the Google analogy dataset in Table \ref{tabResultsGoogle} clearly show that for semantic relations, our model substantially outperforms the two baselines. This is most clear for the \textit{family} relation (e.g.\ ``boy is to girl like brother is to sister''), where we observe an improvement of more than 15 percentage points over \textit{Trans} and almost 36 percentage points over the SVM. This is especially surprising, since the family relation is largely about gender differences, which are normally captured well in word embeddings. For the morphological relations, as expected, our approach was outperformed by the \textit{Trans} baseline. Note that this does not reflect the ability of language models to capture morphological relations, as neural language models are in fact known to be particularly strong in that respect. Rather, this is a consequence of the way in which our templates are obtained, and the fact that morphological relations are typically not explicitly asserted in sentences.

The results for DiffVec in Table \ref{tabResultsDiffVec} follow a similar pattern. For the morphological relations, in this case, our model performs particularly poorly. For instance, for the \textit{Verb 3rd} relation (e.g.\ \textit{accept}-\textit{accepts}), the \textit{Trans} baseline achieves an F1 score of 97.0, compared to only 20.3 for our model. As already mentioned, however, it is not unexpected that our model is unsuitable for such relations. More surprising, perhaps, is the fact that our model also performs poorly on lexical relations such as hypernymy, where the \textit{Trans} baseline achieves an F1 score of 73.6 compared to only 54.3 for our model. For the other types of relations however, our model consistently outperforms the baseline (with the \textit{Expression} relation as the only exception). These other relations are about attributive knowledge, causality, and other forms of commonsense knowledge.

Finally, for BATS (Table \ref{tabResultsBATS}) we see poor performance on morphological relations, mixed performance on lexical relations such as hypernymy and meronymy, and strong results for encyclopedic relations, which is in line with the results we obtained for the other datasets. Note in particular that the model was able to obtain reasonable results for the has-color relation (\textit{thing:color}), which was not possible with the simple hand-coded pattern we used in Table \ref{tabSentencesBERT}.

\subsection{Qualitative analysis}
Finally, we shed some light on the kinds of templates that were identified with our method (see Section \ref{filtering}). Table \ref{templatesqualitative1} shows five templates which were obtained for the \textit{currency} and \textit{capital-of} relations. The first three examples on the right are templates which all explicitly mention the \textit{capital-of} relationship, but they offer more linguistic context than typical manually defined templates, which makes the sentences more natural. In general, we have found that BERT tends to struggle with shorter sentences. There are also patterns that give more implicit evidence of a capital-of relationships, such as the two last ones for the \textit{capital-of} relation. These capture indirect evidence, e.g. the fact that embassies are usually located in the capital of a country.

We also carried out a preliminary comparison with the kind of short manually defined patterns that have been used in previous works.
In our setting, we found the performance of such manually specified templates to be poor, which could suggest that BERT struggles with short sentences. In Table \ref{templatesqualitative2} we show some examples of simple hand-crafted templates in the line of \citet{langmodelknowledgebases2019} and their performance in comparison with our automatically-constructed ones. Note that the results are not strictly comparable as our model uses multiple templates. However, this result does reinforce the importance of using automatic methods to extract templates. Moreover, for many of the diverse relationships which can be found in DiffVec, for example, it can be difficult to come up with meaningful patterns manually. 

\section{Conclusions}
We have studied the question of whether, or to what extent, relational knowledge can be derived from pre-trained language models such as BERT. In particular, we have shown that high-quality relational knowledge can be obtained in a fully automated way, without requiring any hand-coded templates. The main idea is to identify suitable templates using a text corpus, by selecting sentences that mention word pairs which are known to be instances of the considered relation, and then filtering these sentences to identify templates that are predictive of the relation. We have experimentally obtained strong results, although the method is not suitable for all types of relations. In particular, as could be expected, we found that our proposed method is not suitable for morphological relations. More surprisingly, we also found that it performs broadly on par with methods that rely on pre-trained word vectors when it comes to lexical relations such as meronymy and hypernymy. However, for relations that require encyclopedic or commonsense knowledge, we found that our model consistently, and often substantially, outperformed methods relying on word vectors. This shows that the BERT language model indeed captures commonsense and factual knowledge to a greater extent than word vectors, and that such knowledge can be extracted from these models in a fully automated way. 

\smallskip
\noindent\textbf{Acknowledgments.}
Steven Schockaert was funded by ERC Starting Grant 637277.

\bibliography{aaai20}
\bibliographystyle{aaai20}

\end{document}